\documentclass{article}
\usepackage{arxiv}

\publisedin{\scriptsize Preprint of article published in \textit{Neural Information Processing, ICONIP 2019}. 
    See \url{https://doi.org/10.1007/978-3-030-36808-1_50}}
\usepackage{graphicx}
\usepackage{amsmath}

\usepackage{booktabs}

\usepackage{nicefrac}

\usepackage{tikz}
\usetikzlibrary{decorations.pathreplacing, positioning, shapes}%, external}
\tikzset{>=latex}
\usepackage{subfig}

\usepackage{layouts}

\usepackage{hyperref}
\hypersetup
{
    pdftitle = {Deep Learning and Statistical Models for Time-Critical Pedestrian Behaviour Prediction},
    pdfauthor = {Joel Janek Dabrowski},
}

\begin{document}
	%
%	\title{Machine Learning versus Statistical Models for Pedestrian Behaviour Inference}
%	\title{Suitability of Machine Learning and Statistical Models for Pedestrian Behaviour Inference}
	\title{Deep Learning and Statistical Models for Time-Critical Pedestrian Behaviour Prediction}
	\author{
        Joel Janek Dabrowski \\
        Data61, CSIRO, Australia, \\ 
        \texttt{joel.dabrowski@data61.csiro.au}
        \And
		Johan Pieter de Villiers \\
        University of Pretoria, South Africa \\
        CSIR, Pretoria, South Africa \\
        \texttt{pieter.devilliers@up.ac.za}
        \And
		Ashfaqur Rahman \\
        Data61, CSIRO, Australia \\
        \texttt{ashfaqur.rahman@data61.csiro.au}
        \And
		Conrad Beyers \\
        University of Pretoria, South Africa \\
        \texttt{conrad.beyers@up.ac.za}
    }
	\maketitle              % typeset the header of the contribution
	\begin{abstract}
%		objective
%		summary
%		novelty
%		contribution
%		analysis
		The time it takes for a classifier to make an accurate prediction can be crucial in many behaviour recognition problems. For example, an autonomous vehicle should detect hazardous pedestrian behaviour early enough for it to take appropriate measures. In this context, we compare the switching linear dynamical system (SLDS) and a three-layered bi-directional long short-term memory (LSTM) neural network, which are applied to infer pedestrian behaviour from motion tracks. We show that, though the neural network model achieves an accuracy of 80\%, it requires long sequences to achieve this (100 samples or more). The SLDS, has a lower accuracy of 74\%, but it achieves this result with short sequences (10 samples). To our knowledge, such a comparison on sequence length has not been considered in the literature before. The results provide a key intuition of the suitability of the models in time-critical problems.
		
	\end{abstract}

%	\printinunitsof{in}\prntlen{\textwidth}
	
	%Sections
	%% Joel Janek Dabrowski
%% Pedestrian Behaviour Inference
%% Section: Introduction

\section{Introduction}

Many practical applications can be represented as a sequence of various behaviours. These include detecting bull and bear financial markets \cite{maheu2009extracting}, gesture recognition \cite{liu2014fusion}, animal behaviour recognition \cite{leos2017analysis}, and aircraft manoeuvres in military applications \cite{lee2017threat}. Detection time in such applications is often crucial. When data arrives sequentially, detection time becomes a problem of how many sequential samples the model requires to make an accurate prediction.

In this study, we consider the problem of pedestrian behaviour prediction or intent estimation. A review on the prediction of pedestrian behaviour in urban scenarios is presented in \cite{Ridel2018Literature}. A large portion of the literature has been devoted to tracking and path prediction. Many studies use the SLDS as a framework \cite{Schneider2013Pedestrian}, \cite{Kooij2014Analysis}, \cite{kooij2014context}, and \cite{Kooij2016Mixture}. Recently, the recurrent neural network (RNN) has been shown to be a promising approach \cite{hug2018particle,Saleh2018Intent,Saleh2018Long,Cheng2018Pedestrian}. Owing to the significant advancement of the state-of-the-art in pedestrian detection \cite{Li2018Scale,Du2017Deep,Hosang2015Taking,Cai2015Learning}, we assume that the trajectories of the pedestrians are known in this study. Given the pedestrian trajectories, we predict a particular behavioural class.

There are studies have considered the problem of pedestrian behaviour prediction. Probabilistic models such as the latent dynamic conditional random field \cite{Schulz2015Pedestrian} and balanced Gaussian process dynamical models \cite{Minguez2018Pedestrian} have been applied. Various forms of the RNN have been also been considered. Hoy et. al. \cite{Hoy2018Learning} propose a variational RNN which performs both tracking and behaviour prediction. V\"{o}lz et. al. \cite{Volz2016data} compare neural networks, a support vector machine, and the LSTM. 
Though both statistical and machine learning models have been applied to the problem and some studies consider time-to-event analyses, to our knowledge, no specific analyses between these model types in terms of time-to-detection have been considered in the literature.

Our contribution is a comparison between a SLDS and a multi-layered bi-directional LSTM neural network in the context of time-to-detection. This is performed by classifying various pedestrian behaviours from the raw motion tracks under varying sequence lengths. Through the comparison, we gain novel insight into a key difference between the models: though the neural network is more accurate than the SLDS overall, it requires 10 times as many sequential samples to achieve this accuracy. The SLDS is able to provide its most accurate classification within the first few samples of the sequence. This result is important in situations where early detection is imperative.

%	\input{relatedWork.tex}
	%% Joel Janek Dabrowski
%% Pedestrian Behaviour Inference
%% Section: Switching Linear Dynamical System

\section{Switching Linear Dynamical System}

The SLDS models a system that switches between various dynamical models. Each dynamical model is represented as a Linear Dynamic System (LDS) -- which is widely associated with the Kalman filter. The SLDS has been extended in various ways, such as introducing variables representing behavioural context information. Such models have been applied to maritime piracy applications \cite{dabrowski2015Maritime,dabrowski20156unified} and abalone poaching applications \cite{dabrowski2017Context}. Linderman et. al. \cite{Linderman2017Bayesian} extend the SLDS by allowing the switching state to depend on the latent state and exogenous inputs through a logistic regression. The SLDS has also been extended to include multiple LDSs over several sequences under a single switching state \cite{dabrowski2018Naive,dabrowski2016Systemic}.

\begin{figure}[!t]
	\centering
	%Joel Dabrowski
%Attractor distribution
%Figure: Latent dynamic model
%

\def\horisep{1.8}
\def\vertsep{-1.2}

\begin{tikzpicture}
	\tikzstyle{every path}=[->,draw=black!60, thick]
	\tikzstyle{vNode}=[circle,draw=black!60, fill=black!10,minimum size=22pt,inner sep=0pt] 
	\tikzstyle{hNode}=[circle,draw=black!60, minimum size=22pt,inner sep=0pt] 
	\tikzstyle{label}=[];
	%Switchin nodes
	\node[label] (dots) at (-0.4*\horisep,0*\vertsep) {$\cdots$};
	\node[hNode] (stm1) at (0*\horisep,0*\vertsep) {$s_{t-1}$};
	\node[hNode] (st) at (1*\horisep,0*\vertsep) {$s_{t}$};
	\node[hNode] (stp1) at (2*\horisep,0*\vertsep) {$s_{t+1}$};
	\node[label] (dots) at (2.43*\horisep,0*\vertsep) {$\cdots$};
	%Hidden nodes
	\node[label] (dots) at (-0.4*\horisep,1*\vertsep) {$\cdots$};
	\node[hNode] (htm1) at (0*\horisep,1*\vertsep) {$h_{t-1}$};
	\node[hNode] (ht) at (1*\horisep,1*\vertsep) {$h_{t}$};
	\node[hNode] (htp1) at (2*\horisep,1*\vertsep) {$h_{t+1}$};
	\node[label] (dots) at (2.43*\horisep,1*\vertsep) {$\cdots$};
	%Visible nodes
	\node[label] (dots) at (-0.4*\horisep,2*\vertsep) {$\cdots$};
	\node[vNode] (vtm1) at (0*\horisep,2*\vertsep) {$v_{t-1}$};
	\node[vNode] (vt) at (1*\horisep,2*\vertsep) {$v_{t}$};
	\node[vNode] (vtp1) at (2*\horisep,2*\vertsep) {$v_{t+1}$};
	\node[label] (dots) at (2.43*\horisep,2*\vertsep) {$\cdots$};
	
	\draw[->] (stm1) -- (st) node[midway, above] {};
	\draw[->] (st) -- (stp1) node[midway, above] {};
	\draw[->] (stm1) -- (htm1) node[midway, above] {};
	\draw[->] (st) -- (ht) node[midway, right] {};
	\draw[->] (stp1) -- (htp1) node[midway, above] {};
	
	\draw[->] (htm1) -- (ht) node[midway, above] {};
	\draw[->] (ht) -- (htp1) node[midway, above] {};
	\draw[->] (htm1) -- (vtm1) node[midway, above] {};
	\draw[->] (ht) -- (vt) node[midway, right] {};
	\draw[->] (htp1) -- (vtp1) node[midway, above] {};
	
	\draw [-,decorate,decoration={brace, amplitude=5pt}] 
	(2.8*\horisep,0*\vertsep + 0.4) -- (2.8*\horisep,1*\vertsep+0.45)
	node[midway, right, xshift=5pt] {Switching state};
	
	\draw [-,decorate,decoration={brace, amplitude=5pt}] 
	({2.8*\horisep},{1*\vertsep + 0.4}) -- ({2.8*\horisep},{2*\vertsep -0.4})
	node[midway, right, xshift=5pt] {LDS};

\end{tikzpicture}
	\caption{The graphical model of the switching linear dynamical system (SLDS).}
	\label{fig:slds}
\end{figure}
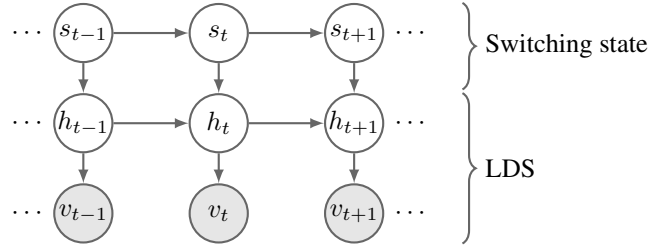

The graphical model representation of the SLDS is illustrated in \figurename~\ref{fig:slds}. The model comprises a switching state variable $s_t$, a hidden or latent variable $h_t$ and a visible or observable variable $v_t$ at time $t$. The latent variable $h_{1:T}$ and observable variable $v_{1:T}$ form a LDS, where the subscript $1:T$ denotes the joint random variable over all discrete time instances 1 to T.. The switching state variable provides the means to switch between various dynamical models. The continuous dynamics of the system are represented by a linear-Gaussian state space model. The following equations describe the system \cite{Barber2012Bayesian,murphy2012machine}
\begin{align}
\label{eq:transition_model}
&h_t = A_t(s_t) h_{t-1} + \eta^h_t(s_t), \\
\label{eq:emission_model}
&v_t = B_t(s_t) h_{t} + \eta^v_t(s_t).
\end{align}
Equation (\ref{eq:transition_model}) describes the transition model and (\ref{eq:emission_model}) describes the emission model. The matrix $A_t$ is the state matrix and $B_t$ is the measurement matrix. With the Gaussian assumption, the noise components are modelled as white noise such that $\eta^h_t(s_t) \sim \mathcal{N}(0,\Sigma_H)$ and $\eta^v_t(s_t) \sim \mathcal{N}(0,\Sigma_V)$. All the LDS model parameters are conditionally dependent on $s_t$ at time $t$. This provides the means to define different dynamic models for each switching state. 

The joint distribution describing the SLDS is given by:
\begin{align}
\label{eq:slds_joint_dist}
p(s_{1:T}, h_{1:T},v_{1:T}) = 
p(s_1)p(h_1) \prod_{t=2}^{T} p(s_t|s_{t-1}) p(h_t|h_{t-1}, s_t) \prod_{t=1}^{T}p(v_t|h_t).
\end{align}
The switching state transition probability $p(s_t|s_{t-1})$ is a discrete distribution. It describes how the model switches between various states. The state transition distribution $p(h_t|h_{t-1}, s_t)$ and emission distribution $p(v_t|h_t)$ are assumed to be Gaussian. These describe the dynamics of the system through the linear state space equations. 

Inference in the SLDS involves inferring the latent variables $s_t$ and $h_t$ given the observations $v_{1:t}$. This is typically performed using filtering and smoothing methods. The filtering operation computes the filtered posterior $p(s_t, h_t | v_{1:t})$. The smoothing operation computes the smoothed posterior  $p(s_t, h_t | v_{1:T})$. Exact inference in the SLDS is intractable \cite{Barber2012Bayesian,murphy2012machine}. Approximate inference algorithms such as the Generalised Pseudo Bayesian (GPB) algorithm \cite{Murphy1998Switching} and the Gaussian Sum Smoothing (GSS) algorithm \cite{Barb2006Expectation} have been developed for the SLDS. In this study, the GPB algorithm is used. 

Parameter learning in the SLDS can be performed using the Expectation Maximisation (EM) algorithm \cite{Murphy1998Switching}. In the expectation step, a smoothing algorithm such as GPB can be used. In the maximisation step, the parameters are estimated using maximum likelihood.

	%% Joel Janek Dabrowski
%% Pedestrian Behaviour Inference
%% Section: Multi-Layered Bidirectional LSTM

\section{Multi-Layered Bidirectional LSTM}

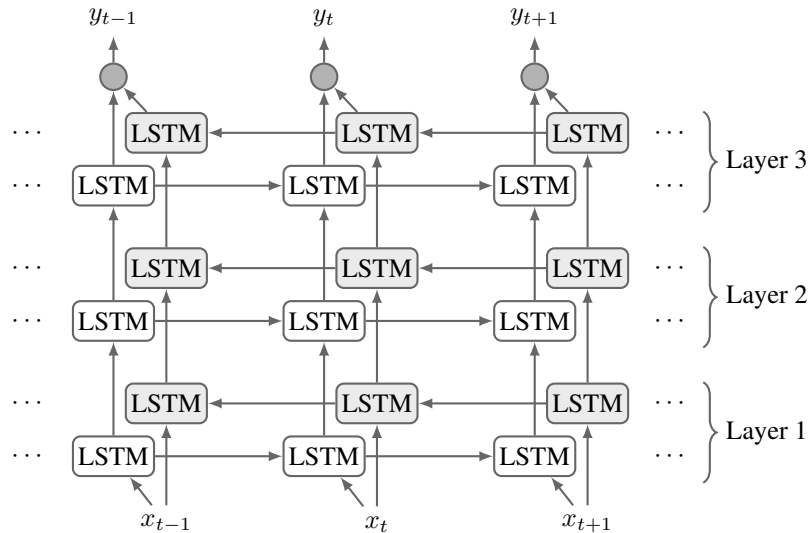
\begin{figure}[!t]
	\centering
	%Joel Dabrowski
%Attractor distribution
%Figure: Latent dynamic model
%

\def\horisep{2.8cm}
\def\vertsep{1.8cm}
\def\offset{20pt}

\begin{tikzpicture}
	\tikzstyle{every path}=[->,draw=black!60, thick]
	\tikzstyle{fcell}=[draw=black!60, fill=black!0, minimum size=15pt, inner sep=2pt, rounded corners=0.1cm] 
	\tikzstyle{bcell}=[draw=black!60, fill=black!8, minimum size=15pt, inner sep=2pt, rounded corners=0.1cm] 
	\tikzstyle{output}=[circle, draw=black!60, fill=black!30, minimum size=10pt, inner sep=0pt] 
	\tikzstyle{label}=[];
	
	\node[label] (in1) at (0*\horisep + 1*\offset,-0.5*\vertsep) {$x_{t-1}$};
	\node[label] (in2) at (1*\horisep + 1*\offset,-0.5*\vertsep) {$x_{t}$};
	\node[label] (in3) at (2*\horisep + 1*\offset,-0.5*\vertsep) {$x_{t+1}$};
	
	%Layer 1
	% Forward
	\node[label] (dots) at (-0.4*\horisep,0*\vertsep) {$\cdots$};
	\node[fcell]%, pin={[pin distance=10pt, pin edge={<-,draw=black!60,thick}]below:$x_{t-1}$}]
	 (fcell11) at (0*\horisep,0*\vertsep) {LSTM};
	\node[fcell]%, pin={[pin distance=10pt, pin edge={<-,draw=black!60,thick}]below:$x_{t}$}]
	 (fcell12) at (1*\horisep,0*\vertsep) {LSTM};
	\node[fcell]%, pin={[pin distance=10pt, pin edge={<-,draw=black!60,thick}]below:$x_{t+1}$}]
	 (fcell13) at (2*\horisep,0*\vertsep) {LSTM};
	\node[label] (dots) at (2.4*\horisep + \offset,0*\vertsep) {$\cdots$};
	% Backward
	\node[label] (dots) at (-0.4*\horisep,0*\vertsep + \offset) {$\cdots$};
	\node[bcell]%, pin={[pin distance=10pt, pin edge={<-,draw=black!60,thick}]below:$x_{t-1}$}]
	(bcell11) at (0*\horisep + \offset,0*\vertsep + \offset) {LSTM};
	\node[bcell]%, pin={[pin distance=10pt, pin edge={<-,draw=black!60,thick}]below:$x_{t}$}]
	(bcell12) at (1*\horisep + \offset,0*\vertsep + \offset) {LSTM};
	\node[bcell]%, pin={[pin distance=10pt, pin edge={<-,draw=black!60,thick}]below:$x_{t+1}$}]
	(bcell13) at (2*\horisep + \offset,0*\vertsep + \offset) {LSTM};
	\node[label] (dots) at (2.4*\horisep + \offset,0*\vertsep + \offset) {$\cdots$};

	% Inputs -> layer 1
	\draw[->] (in1) -- (fcell11);
	\draw[->] (in2) -- (fcell12);
	\draw[->] (in3) -- (fcell13);
	\draw[->] (in1) -- (bcell11);
	\draw[->] (in2) -- (bcell12);
	\draw[->] (in3) -- (bcell13);
	
	% Links between cells
	\draw[->] (fcell11) -- (fcell12);
	\draw[->] (fcell12) -- (fcell13);
	\draw[<-] (bcell11) -- (bcell12);
	\draw[<-] (bcell12) -- (bcell13);
	
	%Layer 2
	% Forward
	\node[label] (dots) at (-0.4*\horisep,1*\vertsep) {$\cdots$};
	\node[fcell] (fcell21) at (0*\horisep,1*\vertsep) {LSTM};
	\node[fcell] (fcell22) at (1*\horisep,1*\vertsep) {LSTM};
	\node[fcell] (fcell23) at (2*\horisep,1*\vertsep) {LSTM};
	\node[label] (dots) at (2.4*\horisep + \offset,1*\vertsep) {$\cdots$};
	% Backward
	\node[label] (dots) at (-0.4*\horisep,1*\vertsep + \offset) {$\cdots$};
	\node[bcell] (bcell21) at (0*\horisep + \offset,1*\vertsep + \offset) {LSTM};
	\node[bcell] (bcell22) at (1*\horisep + \offset,1*\vertsep + \offset) {LSTM};
	\node[bcell] (bcell23) at (2*\horisep + \offset,1*\vertsep + \offset) {LSTM};
	\node[label] (dots) at (2.4*\horisep + \offset,1*\vertsep + \offset) {$\cdots$};
	
	% Layer 1 -> layer 2
	\draw[->] (fcell11) -- (fcell21);
	\draw[->] (fcell12) -- (fcell22);
	\draw[->] (fcell13) -- (fcell23);
	\draw[->] (bcell11) -- (bcell21);
	\draw[->] (bcell12) -- (bcell22);
	\draw[->] (bcell13) -- (bcell23);
	
	% Links between cells
	\draw[->] (fcell21) -- (fcell22);
	\draw[->] (fcell22) -- (fcell23);
	\draw[<-] (bcell21) -- (bcell22);
	\draw[<-] (bcell22) -- (bcell23);
	
	%Layer 3
	% Forward
	\node[label] (dots) at (-0.4*\horisep,2*\vertsep) {$\cdots$};
	\node[fcell] (fcell31) at (0*\horisep,2*\vertsep) {LSTM};
	\node[fcell] (fcell32) at (1*\horisep,2*\vertsep) {LSTM};
	\node[fcell] (fcell33) at (2*\horisep,2*\vertsep) {LSTM};
	\node[label] (dots) at (2.4*\horisep + \offset,2*\vertsep) {$\cdots$};
	% Backward
	\node[label] (dots) at (-0.4*\horisep, 2*\vertsep + \offset) {$\cdots$};
	\node[bcell] (bcell31) at (0*\horisep + \offset,2*\vertsep + \offset) {LSTM};
	\node[bcell] (bcell32) at (1*\horisep + \offset,2*\vertsep + \offset) {LSTM};
	\node[bcell] (bcell33) at (2*\horisep + \offset,2*\vertsep + \offset) {LSTM};
	\node[label] (dots) at (2.4*\horisep + \offset,2*\vertsep + \offset) {$\cdots$};
	
	% Layer 1 -> layer 2
	\draw[->] (fcell21) -- (fcell31);
	\draw[->] (fcell22) -- (fcell32);
	\draw[->] (fcell23) -- (fcell33);
	\draw[->] (bcell21) -- (bcell31);
	\draw[->] (bcell22) -- (bcell32);
	\draw[->] (bcell23) -- (bcell33);
	
	% Links between cells
	\draw[->] (fcell31) -- (fcell32);
	\draw[->] (fcell32) -- (fcell33);
	\draw[<-] (bcell31) -- (bcell32);
	\draw[<-] (bcell32) -- (bcell33);

	%Output layer
	\node[output, pin={[pin distance=10pt, pin edge={->,draw=black!60,thick}]above:$y_{t-1}$}]
	(cell31) at (0*\horisep,3*\vertsep - 0.5*\offset) {};
	\node[output, pin={[pin distance=10pt, pin edge={->,draw=black!60,thick}]above:$y_{t}$}]
	(cell32) at (1*\horisep,3*\vertsep - 0.5*\offset) {};
	\node[output, pin={[pin distance=10pt, pin edge={->,draw=black!60,thick}]above:$y_{t+1}$}]
	(cell33) at (2*\horisep,3*\vertsep - 0.5*\offset) {};
	
	% Layer 2 -> Output layer
	\draw[->] (fcell31) -- (cell31);
	\draw[->] (fcell32) -- (cell32);
	\draw[->] (fcell33) -- (cell33);
	\draw[->] (bcell31) -- (cell31);
	\draw[->] (bcell32) -- (cell32);
	\draw[->] (bcell33) -- (cell33);
	
	\draw [-,decorate,decoration={brace, amplitude=5pt}] 
	(2.8*\horisep, 0*\vertsep + 1.4*\offset) -- (2.8*\horisep, 0*\vertsep - 0.5*\offset)
	node[midway, right, xshift=5pt] {Layer 1};
	
	\draw [-,decorate,decoration={brace, amplitude=5pt}] 
	(2.8*\horisep, 1*\vertsep + 1.4*\offset) -- (2.8*\horisep, 1*\vertsep - 0.5*\offset)
	node[midway, right, xshift=5pt] {Layer 2};
	
	\draw [-,decorate,decoration={brace, amplitude=5pt}] 
	(2.8*\horisep, 2*\vertsep + 1.4*\offset) -- (2.8*\horisep, 2*\vertsep - 0.5*\offset)
	node[midway, right, xshift=5pt] {Layer 3};
	
\end{tikzpicture}
	\caption{Three-layered bi-directional LSTM architecture. Each rectangular node denotes an LSTM cell. Round nodes denote softmax layers. The edges denote connectivity between the LSTM cells and output layer. At time $t$, pedestrian tracks are denoted by $x_t$ and the behaviour class is denoted by $y_t$.}
	\label{fig:rnn}
\end{figure}

A three-layered bi-directional LSTM \cite{hochreiter1997long} RNN is constructed for comparison with the SLDS. The model is illustrated in \figurename{~\ref{fig:rnn}}. Each LSTM layer comprises two sequences of LSTM cells; one propagating in the positive time direction and one in the negative time direction. Together, the forward and backward sequences form a bi-directional LSTM (BiLSTM). The bi-directional structure provides a means to make a prediction at time $t$ according to the full sequence $1:T$, $t \leq T$. This is in comparison to a uni-directional structure, which makes a prediction at time $t$ according to the sequence $1:t$. Three BiLSTMs are stacked to form three distinct layers. Multiple layers provide a deep structure which promotes higher level feature extraction. Input data is provided to the inputs of the first BiLSTM layer. For each sequence step, the outputs of the third BiLSTM layer are passed through a softmax layer. The softmax outputs the predicted class associated with the current input sample. For notational simplicity, this model is referred to as the RNN in the remainder of the discussion.
%

	%% Joel Janek Dabrowski
%% Pedestrian Behaviour Inference
%% Section: Dataset

\section{Dataset}

The well-known Daimler Pedestrian Path Prediction Benchmark Dataset (GCPR’13) \cite{Schneider2013Pedestrian} is used in this study. The dataset comprises a collection of 68 pedestrian sequences with 4 different pedestrian behaviour types: crossing, stopping, starting to walk, and bending-in. Though the dataset seems relatively small, the LSTM has been shown to perform well in the path prediction application \cite{Saleh2018Intent}. 

The dataset was acquired using stereo cameras. The stereo camera provides the means to produce three dimensional Cartesian coordinates for tracking purposes. The ground truth of the dataset provides bounding boxes, disparity, and $X$ and $Z$ coordinates of each target.

The dataset was developed to test recursive Bayesian filters for pedestrian path prediction for different behaviours \cite{Schneider2013Pedestrian}. In this study, this problem is inverted. The behaviour of a pedestrian is inferred from the tracked path.

	%% Joel Janek Dabrowski
%% Pedestrian Behaviour Inference
%% Section: Approach

\section{Methodology}

The dataset is provided with a predefined training set and a test set. The training and test sets comprise 36 and 32 sequences respectively. The model parameters are estimated using the training dataset. The trained models are then applied to predict the behaviour class from the tracks provided in the test dataset. To measure the performance of the models, accuracy, precision and recall are used.

The models are tested on sequences of varying length. This is achieved by truncating the sequences in increments of 10 samples. That is, the models are tested on the first $10, 20, 30, \dots$ samples of each sequence in the test set. The classification results are stored for each sequence length. Limiting the number of timesteps provides an indication of how well the method is able to predict a behaviour class in a short period of time. Furthermore, it provides some form of consistency over the varying sequence lengths in the dataset.

The SLDS motion model is configured as a constant acceleration model. The tracked coordinates are provided as observations to the SLDS. The model parameters are learned using the EM algorithm. The switching state is defined to comprise the states BendingIn, Crossing, Starting, and Stopping. In the dataset, the pedestrians do not switch between behaviour classes over the sequences. The switching state transition distribution is set with a $0.97$ probability of remaining in the current switching state and a $0.01$ probability of transitioning to one of the other three switching states. The prior switching state probability distribution is set to the uniform distribution.

The RNN is configured with 32 hidden units in each LSTM cell. The ADAM algorithm \cite{kingma2014adam} is used to minimise the cross entropy of the softmax outputs. The model is trained over 110 epochs with a learning rate of 0.0001 and a batch size of 1. The remaining ADAM parameters are set as recommended in \cite{kingma2014adam}. The RNN is trained over the complete length of each sequence in the test set.
	%% Joel Janek Dabrowski
%% Pedestrian Behaviour Inference
%% Section: Results

\section{Results}

%
% See plotResults.py
\begin{figure}[!t]
	\centering
	\includegraphics{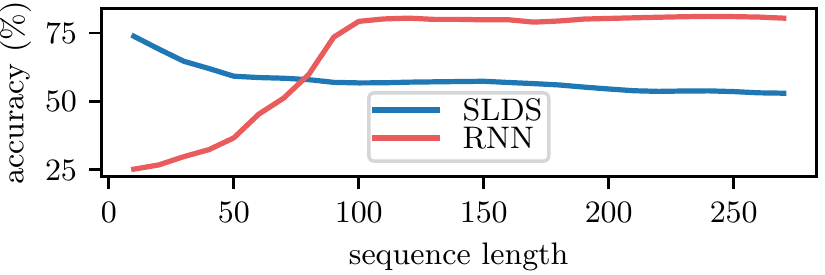}
	\caption{Accuracy over the set of truncated sequences.}
	\label{fig:accuracy}
\end{figure}
The accuracy over the set of truncated sequences is presented in \figurename{~\ref{fig:accuracy}}. The striking feature in the plot is that the RNN increases in accuracy with increasing sequence length, whereas the SLDS decreases in accuracy with increasing sequence length. The SLDS has the highest accuracy with a sequence of 10 samples. This implies that within the first 10 samples, the SLDS is able to classify the sequence. The RNN's accuracy curve saturates around the 100 sample length mark. This indicates that the RNN requires a sequence of at least 100 samples to achieve the high accuracy.

These results are consistent with theoretical design of the models. The SLDS assumes a first order Markov model in both the dynamics and the switching state. A first order Markov model assumes that the current state is conditionally dependent \textit{only} on the previous state. The result is that the SLDS is not designed to model long-term dependencies in the data. The SLDS thus performs better when provided with the shorter sequences. Furthermore, the SLDS performance decreases with sequence length as it is designed to switch between dynamics. It is more likely to switch behaviour class in a longer sequence. The LSTM cell in the RNN has been specifically designed to model both long and short-term dependencies in the data \cite{hochreiter1997long}. The result is that the RNN requires a longer sequence to achieve a higher accuracy. Another relevant difference between the models is that the SLDS is a structured model where the dynamics have been predefined. In the RNN, the dynamics are learned in a black-box approach, which often requires more data.

\begin{figure}[!t]
	\centering
	\includegraphics{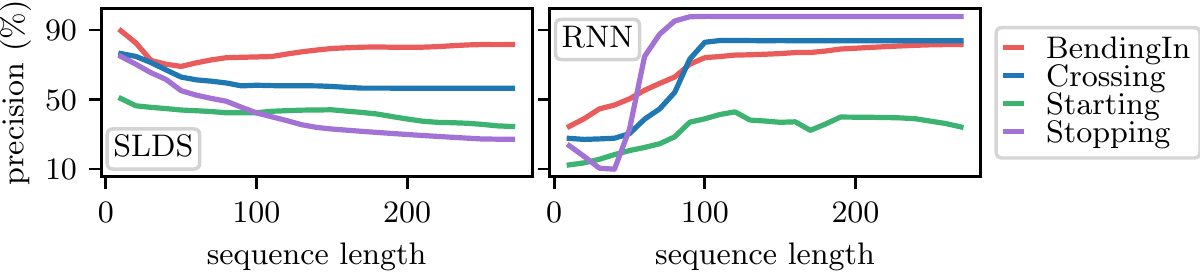}
	\caption{Precision over the set of truncated sequences.}
	\label{fig:precision}
\end{figure}

\begin{figure}[!t]
	\centering
	\includegraphics[width=4.8in]{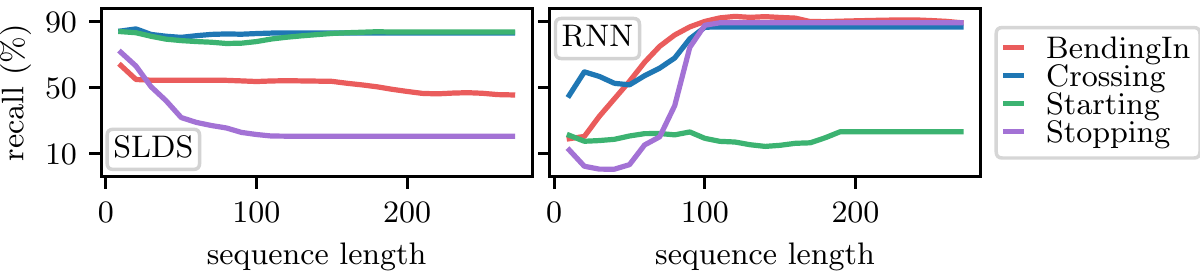}
	\caption{Recall over the set of truncated sequences.}
	\label{fig:recall}
\end{figure}

The precision and recall over the set of truncated sequences are presented in \figurename{~\ref{fig:precision}} and \figurename{~\ref{fig:recall}} respectively. Confusion matrices for the 10-sample-length and complete sequences are presented in Table \ref{table:confusionMatrices}. Precision is often viewed as a measure of the quality of the model. Recall describes the probability of correctly classifying the pedestrian behaviour.

As for the RNN accuracy, the precision and recall values are only high for sequences with 100 samples or more. The precision and recall for the SLDS are highest for sequences of 10 steps.

The RNN generally has a higher precision and recall than the SLDS. The RNN however struggles to correctly predict the starting behaviour class. Consider the confusion matrix for the complete sequence classification presented in Table \ref{table:confusionMatrices}. The majority of Starting samples are incorrectly associated with the BendingIn class. A possible reason for this is that the sequences for the Starting class are generally short in length. The poor results for the Starting class lowers the overall accuracy of the RNN.

%
% See plotResults.py
%
%
% See plotResults.py
%
%
% See plotResults.py
\begin{table}[!t]
	\centering
	\caption{Confusion matrices for the SLDS and RNN for the 10-sample-length and complete sequence predictions. The matrices are normalised over the rows to indicate a form of recall. Rows and columns follow the class order of BendingIn, Crossing, Starting, and Stopping.}
	\setlength{\tabcolsep}{10pt}
	\begin{tabular}{c c c}
		\toprule
		& 10 samples & Complete sequence \\
		\midrule
		SLDS &
		$
		\begin{bmatrix}
		0.63 & 0.02 & 0.16 & 0.00 \\
		0.00 & 0.84 & 0.00 & 0.29 \\
		0.30 & 0.00 & 0.84 & 0.00 \\
		0.07 & 0.13 & 0.00 & 0.71 \\
		\end{bmatrix}
		$ & 
		$
		\begin{bmatrix}
		0.46 & 0.04 & 0.04 & 0.23 \\
		0.02 & 0.83 & 0.12 & 0.51 \\
		0.37 & 0.01 & 0.84 & 0.06 \\
		0.15 & 0.12 & 0.00 & 0.21 \\
		\end{bmatrix}
		$ \\
		\midrule
		RNN & 
		$\begin{bmatrix}
		0.19 & 0.19 & 0.19 & 0.19 \\
		0.46 & 0.45 & 0.47 & 0.47 \\
		0.21 & 0.22 & 0.21 & 0.22 \\
		0.13 & 0.14 & 0.13 & 0.12 \\
		\end{bmatrix}$ & 
		$
		\begin{bmatrix}
		0.89 & 0.10 & 0.67 & 0.00 \\
		0.00 & 0.87 & 0.09 & 0.10 \\
		0.11 & 0.02 & 0.23 & 0.00 \\
		0.00 & 0.02 & 0.00 & 0.89 \\
		\end{bmatrix}
		$ \\
		\bottomrule
	\end{tabular}
	\label{table:confusionMatrices}
\end{table}

The lowest recall is for the SLDS model is the BendingIn class, with a value of $63\%$. Considering the confusion matrix, the $30\%$ of the samples were misclassified as starting behaviour. The model performs well on the crossing and starting classes. For longer sequences, the precision and recall for the Stopping class decreases significantly. As also indicated in \figurename{~\ref{fig:recall}}, the recall for the Crossing and Starting classes remain fairly constant.

For the RNN with 10-sample sequences, $46\%$ of the BendingIn samples were incorrectly associated with the Crossing class as indicated in Table \ref{table:confusionMatrices}. When provided with the complete sequence, this reduces to $0\%$. Similarly, most of the Starting samples are incorrectly associated with the Crossing class with short sequences. When provided with the complete sequence, the incorrect classifications shift to the BendingIn class.

\begin{figure}[!t]
	\centering
	\subfloat[Pedestrian track..]
	{
		\label{fig:track_0_BendingIn}
		\includegraphics{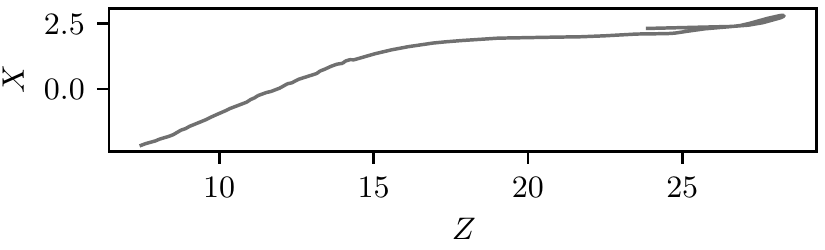}
	} \\
	\subfloat[Behaviour prediction. (Horizontal axis: sequence samples).]
	{
		\label{fig:inference_0_BendingIn}
		\includegraphics[scale=1]{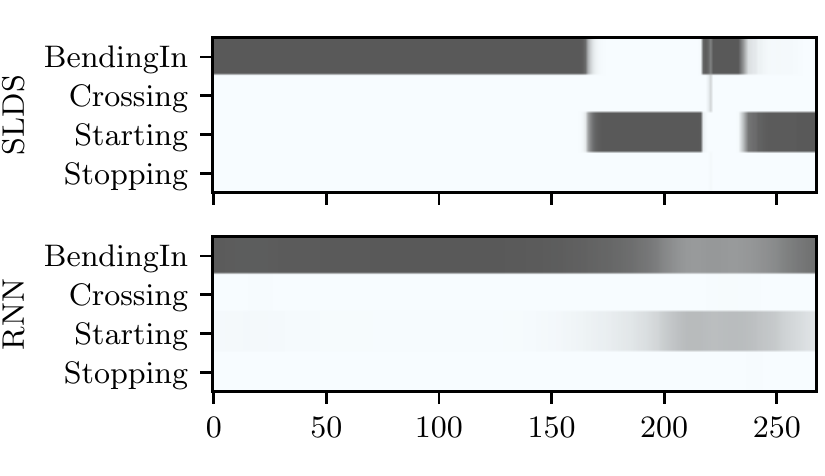}
	}
	\caption{Pedestrian track and behaviour prediction for test sequence 0. The true behaviour class is BendingIn.}
	\label{fig:sequence_0_BendingIn}
\end{figure}

A plot of the track and the class predictions for test sequence 0 is presented in \figurename{~\ref{fig:sequence_0_BendingIn}}. The track of the pedestrian performing the BendingIn activity is presented in \figurename{~\ref{fig:track_0_BendingIn}}. The horizontal axis represents the depth dimension, $Z$ with respect to the camera. The vertical axis represents the camera's horizontal axis, $X$. Note that the time aspect of the track is not represented in this plot. The plot of the predicted switching state over time is presented in \figurename{~\ref{fig:inference_0_BendingIn}}. Dark grey indicates a high probability of that the pedestrian belongs to a particular class. Light grey indicates a low probability of that the pedestrian behaviour belongs to the particular class. Both the SLDS and the RNN associate the behaviour with the BendingIn class for the first 160 time steps. The predictions subsequently transition to the Starting class. This may be explained by the fact that the pedestrian seems to back-track as illustrated in \figurename{~\ref{fig:track_0_BendingIn}}.

\begin{figure}[!t]
	\centering
	\includegraphics{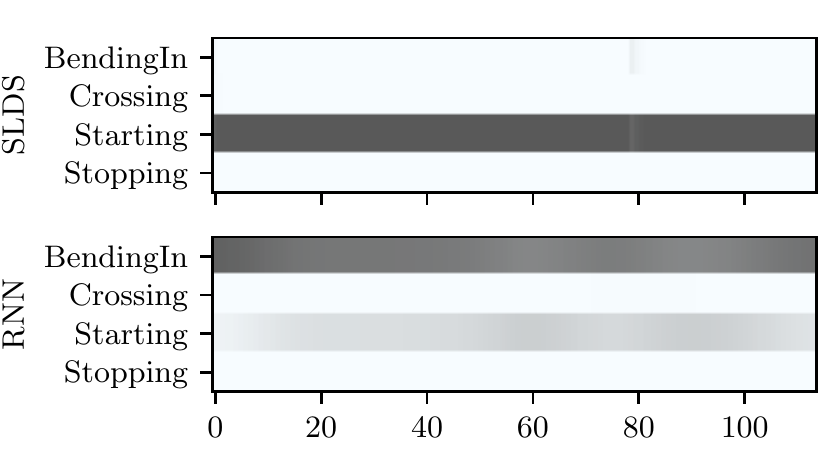}
	\caption{Behaviour prediction for test sequence 11. The true behaviour class is Starting.}
	\label{fig:sequence11_Starting}
\end{figure}

\figurename{~\ref{fig:sequence11_Starting}} illustrates an example of the Starting behaviour class. The SLDS successfully predicts the correct class for the entire sequence. The RNN incorrectly predicts the BendingIn class, but does associate some probability with the Starting class. This result corresponds the complete-sequence confusion matrix presented in Table \ref{table:confusionMatrices}. 

\begin{figure}[t]
	\centering
	\subfloat[Pedestrian track.]
	{
		\label{fig:inference_8_Crossing}
		\includegraphics{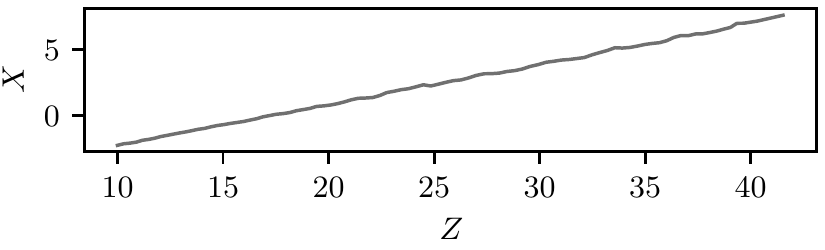}
	} \\
	\subfloat[Behaviour prediction. (Horizontal axis: sequence samples)]
	{
		\label{fig:track_8_Crossing}
		\includegraphics{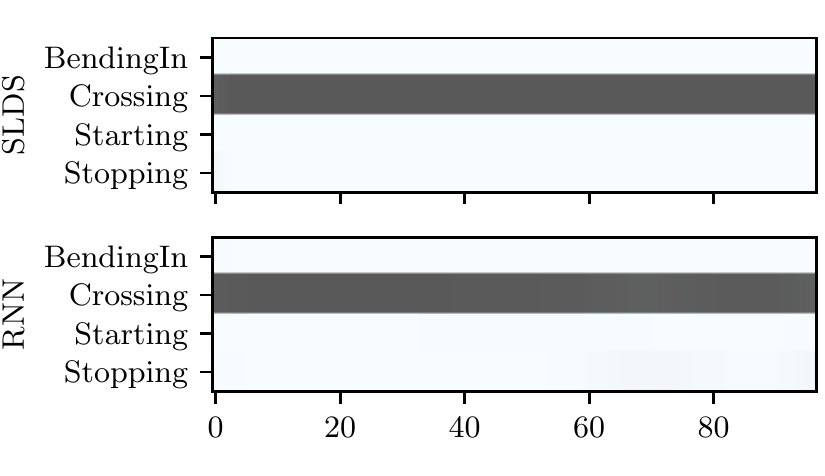}
	}
	\caption{Pedestrian track and behaviour prediction for test sequence 12. The true behaviour class is `crossing'.}
	\label{fig:sequence8_Crossing}
\end{figure}

\figurename{~\ref{fig:sequence8_Crossing}} illustrates results for the Crossing behaviour class. The track in \figurename{~\ref{fig:track_8_Crossing}} is approximately linear over the space. With such behaviour, both models generally perform well in this class.

\begin{figure}[!t]
	\centering
	\includegraphics{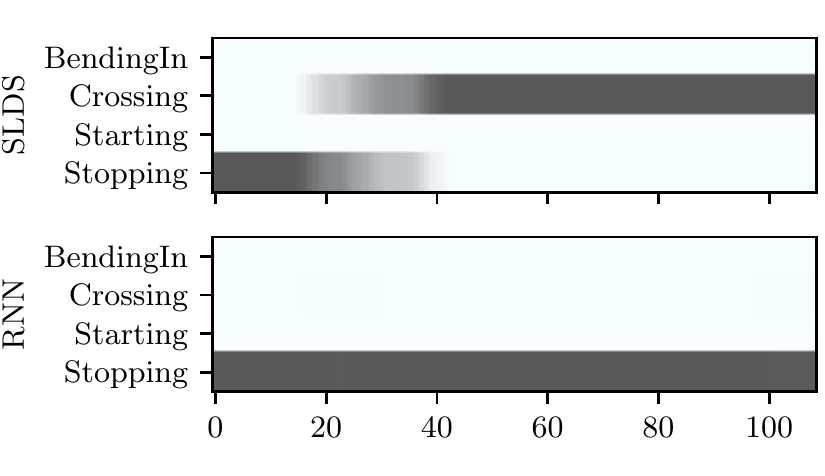}
	\caption{Behaviour prediction for test sequence 11. The true behaviour class is Stopping.}
	\label{fig:sequence9_Stopping}
\end{figure}

An example of the Stopping behaviour class is presented in \figurename{~\ref{fig:sequence9_Stopping}}. The SLDS correctly begins by classifying the stopping behaviour class and then transitions to the crossing class. This result corresponds to the complete sequence confusion matrix presented in Table \ref{table:confusionMatrices}. The RNN correctly classifies the stopping class for the entire sequence. This corresponds to the high recall for this class as illustrated in \figurename{~\ref{fig:recall}}.

	%% Joel Janek Dabrowski
%% Pedestrian Behaviour Inference
%% Section: Conclusion

\section{Summary and Conclusion}

In this study a SLDS and a three-layered bidirectional LSTM RNN are applied to predict pedestrian behaviour from motion tracks from the Daimler Pedestrian Path Prediction Benchmark Dataset (GCPR’13). The key result is that the RNN model's accuracy increases with increasing sequence length, whereas the SLDS's accuracy decreases with increasing sequence length. The best results for the SLDS are obtained when the first 10 samples of the sequence are provided to the model. This is possibly due to the SLDS being designed to model short-term behaviour as well as having a predefined model of the dynamics. The RNN is designed to model both short and long-term dynamics with a black-box approach. The result is that the RNN is more accurate, but over longer sequences (100 samples or more). This suggests that in situations where a decision is required to be made quickly, the SLDS may be the preferred model.

There is potential for improvement of the results for both models. One approach would be to include contextual information. This can be achieved in the SLDS using methods such as those described in \cite{dabrowski2015Maritime,dabrowski20156unified,dabrowski2017Context}. Contextual information could include road signs, proximity to crossing areas, and traffic congestion levels. Additional information relating to the urban environment could also be influential. For example, a street may be residential or commercial.

	\bibliographystyle{unsrt}
	\bibliography{Bibliography}
	
\end{document}